\documentclass{article}

\PassOptionsToPackage{numbers, compress}{natbib}
\usepackage[final]{neurips_2021_ml4ad}

\usepackage[utf8]{inputenc} % allow utf-8 input
\usepackage[T1]{fontenc}    % use 8-bit T1 fonts
\usepackage{hyperref}       % hyperlinks
\usepackage{url}            % simple URL typesetting
\usepackage{booktabs}       % professional-quality tables
\usepackage{nicefrac}       % compact symbols for 1/2, etc.
\usepackage{microtype}      % microtypography
\usepackage{xcolor}
\usepackage{color} 
\usepackage{amsmath,amssymb,amsfonts}
\usepackage{algorithm, algorithmic}
\usepackage{graphicx}
\usepackage{textcomp}
\usepackage{relsize}
\usepackage{multirow}
\usepackage{makecell}
\usepackage{bm}
\usepackage{amsmath}
\DeclareMathOperator*{\argmax}{argmax}
\DeclareMathOperator*{\argmin}{argmin}
\usepackage{subcaption}
\title{Offline Reinforcement Learning for Autonomous Driving with Safety and Exploration Enhancement}

\author{Tianyu Shi \\
   Department of Civil \& Mineral Engineering\\
   University of Toronto\\
  \texttt{ty.shi@mail.utoronto.ca} \\
   \And
    Dong Chen \\
    Department of Mechanical Engineering \\
    Michigan State University \\
   \texttt{chendon9@msu.edu} \\
   \AND
   Kaian Chen \\
   Department of Mechanical Engineering \\
    Michigan State University \\
   \texttt{chenkaia@msu.edu} \\
   \And
   Zhaojian Li \\
   Department of Mechanical Engineering \\
   Michigan State University \\
   \texttt{lizhaoj1@msu.edu} \\
}

\begin{document}

\maketitle

\begin{abstract}
Reinforcement learning (RL) is a powerful data-driven control method that has been largely explored in autonomous driving tasks. However, conventional RL approaches learn control policies through trial-and-error interactions with the environment and therefore may cause disastrous consequences such as collisions when testing in real-world traffic. Offline RL has recently emerged as a promising framework to learn effective policies from previously-collected, static datasets without the requirement of active interactions, making it especially appealing for autonomous driving applications. Despite promising, existing offline RL algorithms such as Batch-Constrained deep Q-learning (BCQ) generally lead to rather conservative policies with limited exploration efficiency. To address such issues, this paper presents an enhanced BCQ algorithm by employing a learnable parameter noise scheme in the perturbation model to increase the diversity of observed actions. In addition, a Lyapunov-based safety enhancement strategy is incorporated to constrain the explorable state space within a safe region. Experimental results in highway and parking traffic scenarios show that our approach outperforms the conventional RL method, as well as  state-of-the-art offline RL algorithms.
\end{abstract}

\section{Introduction}
Autonomous driving has received exceedingly high research interests in the past two decades as it offers the promise of releasing drivers from exhausting driving. While great advances have been achieved in the field of path planning, perception and controls, high-level decision-making remains a challenge especially in mixed traffic with complex and dynamic driving environment. Recently, numerous reinforcement learning (RL) approaches have been applied to autonomous driving tasks and promising results are reported \cite{sallab2017deep, wang2018automated, shi2019driving, chen2020autonomous, chen2021interpretable, wu2021human}. However, conventional RL algorithms evolve through interacting with the environment, via sometimes trial-and-error exploratory actions that make the vehicles vulnerable to accidents in real-world traffic.

Offline RL (also known as batch RL) has been proposed as a promising framework to address the safety issue where agents learn from  pre-collected datasets without interacting with the real-world environment. As such, it has received increased interests in safety-critical applications such as decision making in healthcare, robotics, and autonomous driving \cite{levine2020offline}. In particular, the batch-constrained RL (BCQ) algorithm is proposed in \cite{fujimoto2019off}, where a state-dependent generative model is used to restrict predicted actions to be similar to previously observed ones  to tackle the issue of extrapolation error caused by erroneously estimating seen state-action pairs. In addition, the authors in \cite{wu2019behavior} exploit the schemes of  value penalty factor and  policy regularization in the value and policy objective functions to regularize the learned policy towards the expert policy and worthy performance gains  on recently proposed offline RL methods are obtained. The aforementioned behavior-constrained approaches essentially restrict the learned policy distribution to resemble the datasets to mitigate the effects of extrapolation error, which on the other hand will generally drive the agents to act conservatively  without efficiently exploring the state and action space \cite{fujimoto2019off}. This tends to result in poor diversity of seen state-action pairs, which negatively impairs the learning performance.

Learning to explore is an emerging paradigm to address the issue of insufficient exploration \cite{fujimoto2019off, lillicrap2015continuous, plappert2017parameter, xu2018learning}. For instance,  \cite{plappert2017parameter} has shown improved exploratory behavior through adding additive Gaussian noise to the parameter vectors on 3 off-policy deep RL algorithms. Deep Deterministic Policy Gradient (DDPG) \cite{lillicrap2015continuous} is then used to independently train an exploration policy  by integrating it with an auto-correlated noise added to the actor policy. Despite promising results, the aforementioned approaches apply state-independent noises to enhance exploration, which may not adapt satisfactorily to more diverse environments like the case in autonomous driving. 

In this paper, we build upon the state-of-the-art offline RL algorithm, BCQ, and develop a more efficient RL framework with a learnable parameter noise in the perturbation model to enhance exploration and achieve increased diversity in seen actions. Furthermore, Lyapunov-based safety regulation is adopted to enhance the safety in explorations. The main contributions and the technical advancements of this paper are summarized as follows.
\begin{enumerate}
\item We build upon BCQ and develop a more efficient and safety-enhanced offline RL framework that are applicable to many safety-critical real-world applications. 
\item  A novel learnable parameter noise scheme is employed to enhance the diversity of seen actions and a Lyapunov-based risk factor is constructed to restrict the exploratory state space within the safe region.
\item We conduct comprehensive experiments on autonomous driving in both highway and parking traffic scenarios, and the results show that our approach consistently outperforms standard RL and several state-of-the-art offline RL algorithms in terms of driving safety and efficiency.
\end{enumerate}

The remainder of this paper is organized as follows. Section~\ref{sec:background} briefly introduces the preliminaries of RL, offline RL and Lyapunov stability theory. The proposed offline RL framework with enhanced safety and exploration efficiency is described in Section~\ref{sec:method} whereas experiments, results, and discussions are presented in Section~\ref{sec:exp}. Finally, we conclude the paper and discuss future works in Section~\ref{sec:conclu}.

\section{Background}
\label{sec:background}

\subsection{Preliminaries of Reinforcement Learning}
In a RL setting, the objective is to learn an optimal policy $\pi^{*}$ that maximizes the accumulated return $R = \sum_{t=0}^{T} \gamma ^t r_{t}$, where $r_{t}$ is the reward at time step $t$ and $\gamma\in (0,1)$ is the discount factor. More specifically,  the agent observes the state $s_t \in \mathcal{S} \subseteq \mathbb{R}^{n}$ of the environment at each time $t$, and interacts with the environment by performing an action $a_t \in \mathcal{A} \subseteq \mathbb{R}^{m}$ according to a policy $\pi (a|s)$. The state-action value function (or Q-function) $Q^{\pi}(s_t,a_t)$ of a policy $\pi$ is the expected return when following the policy after taking action $a_t$ in state $s_t$.
The optimal value function $Q^{*}(s_t, a_t)$, representing the reward of taking action $a_t$ in state $s_t$ followed by the optimal policy  $\pi^{*}$ through greedy action choices, can be obtained from the following Bellman equation: 
\begin{equation}
\mathcal{T} Q^{*}(s_t, a_t) =  E [r(s_t, a_t) + \gamma  \sum_{s_{t+1}} \mathcal{P}(s_{t+1} |s_t, a_t) \max_{a_{t+1}} Q^{*}(s_{t+1},a_{t+1})],
\end{equation}
where $\mathcal{T}$ denotes the Bellman operator and $\mathcal{P}(s_{t+1} |s_t, a_t)$ is the transition probability. Off-policy algorithms like Q-learning \cite{sutton2018reinforcement, mnih2013playing} fit the Q-function with a parametric model $Q_{\theta}$ and update the parameters with sampled data from the experience buffer dataset $\mathcal{D} = (s_t, a_t, r_t, s_{t+1})$ \cite{lin1992self}. Actor-critic networks like DDPG \cite{lillicrap2015continuous} adopt two networks: an actor network $\pi_{\theta} (s)$ for policy learning and a critic network $Q_{\phi} (s, a)$ to reduce variance, where the policy network is updated as:
\begin{equation}
\label{eqn:ddpg}
\phi \leftarrow \argmax_{\phi} E_{s \in \mathcal{D}} [Q_{\theta}(s, \pi_{\phi}(s))].
\end{equation}

\subsection{Offline Reinforcement Learning}
Offline Reinforcement learning is essentially a type of off-policy RL that works on a pre-collected and static dataset $\mathcal{B}$ without the requirement of continuous interactions with the environment \cite{levine2020offline, ardoinextracting}. Typically, the dataset $\mathcal{B} = \{(s_t, a_t, r_t, s_{t+1})\}_{t=0:N}$ of unknown quality is first obtained. Batch-Constrained deep Q-learning (BCQ) \cite{fujimoto2019off} is a state-of-the-art offline RL method aiming at enforcing the learned policy to be similar to the behavior policy exhibited in the data. 
BCQ aims to solve a key challenge in offline RL that the values of the seen state-action pairs are often erroneously estimated (also known the as extrapolation error phenomenon). Towards that end, BCQ samples multi-step actions from a generative model (i.e., VAE \cite{kingma2013auto}), which is then used to train the policy by producing actions similar to the ones in the observed data batch: 
\begin{equation}
\label{eqn:bcq}
\pi(s) = \argmax_{\hat{a_i}} Q_{\theta} (s, \hat{a_i}),
\end{equation}
where  $\hat{a}_i = a^n_i + \xi_{\phi} (s, a^n_i)$ with $a^n_i$ being the action generated from a generative model and $\xi_{\phi} (s, a^n_i)$ being a perturbation model added to increase the diversity of seen actions \cite{fujimoto2019off}.  The perturbation model $\xi_{\phi}$ is updated as:
\begin{equation}
\label{eqn:bcqxi}
\phi \leftarrow \argmax_{\phi} \sum_{(s, a) \in \mathcal{B}} Q_{\phi} (s, a),
\end{equation}
and the critic network $Q_{\theta}$ is updated as:
\begin{equation}
\label{eqn:target}
\theta \leftarrow \argmin_{\theta} \sum_{(s, a) \in \mathcal{B}} (y - Q_{\theta} (s, a)),
\end{equation}
where $y$ is a combination of the two target Q-values, $Q_{\theta_1'}$ and $Q_{\theta_2'}$,  from the target networks and is defined as:
\begin{equation}
\label{eqn:bcqxi}
y = r + \gamma \max_{\hat{a_i}} [\lambda \min_{j=1,2} Q_{{\theta_j}'}(s', \hat{a_i}) + (1- \lambda) \max_{j=1,2} Q_{\theta_j'}(s', \hat{a_i})],
\end{equation}
where $\lambda$ is a parameter that controls the uncertainty introduced from future time steps.

\subsection{Lyapunov Stability}\label{subsec:LyapStab}
Consider the following dynamical system:
\begin{equation}\label{equ:state_Trans}
\frac{dx}{dt}=f(x,u)\equiv f_u(x),
\end{equation}
where $x(t)\in D\subseteq \mathbb{R}^n$ is the state vector with $D$ being the domain, and $u(t)\in \mathbb{R}^m$ is the control input vector. The closed-loop system is stable at the origin if for any $\epsilon \in R^+$, there exists $\delta(\epsilon)\in R^+$, such that if $||x(0)||\leq\delta$ then $||x(t)||\leq\epsilon$ for all $t\geq 0$. Furthermore, the system is asymptotically stable if it is stable and the state goes to zero asymptotically, i.e., $\lim_{t\rightarrow \infty}||x(t)||=0$ for all $||x(0)||<\delta$ \cite{chang2019neural}.

Lyapunov theory \cite{Khalil2002NLsys} is a well-studied method to characterize the stability conditions. Specifically, if there exists a continuously differentiable function $V:\mathbb{R}^n\rightarrow\mathbb{R}^+$ for the closed-loop system $f_u(x)$ such that
\begin{equation}
V(0)=0, \text{ and}, V(x) > 0,  \forall x \in D, x\neq0, \text {and } \nabla_{f_u}V(x)<0.
\label{lf criteria}
\end{equation}	
Here $\nabla_{f_u}V(x)$ is the Lie derivative and defined as
\begin{equation}
 \nabla_{f_u}V(x)=\sum^{n}_{i=1} \frac{\partial V}{\partial x_i} \frac{dx_i}{dt}=\sum^{n}_{i=1} \frac{\partial V}{\partial x_i} [f_u]_i (x).
\end{equation}

\section{Methodology}
\label{sec:method}
\subsection{Learning to Explore}
\label{sec:noi}
In BCQ,  a perturbation model $\xi_\phi (s, a)$ parameterized by $\phi$ is used to generate a noise signal, which is added to the VAE-generated action $a$ to facilitate exploration and increase the diversity of the seen actions. As reported in \cite{plappert2017parameter},  injecting parameter noises within traditional RL methods can generally promote the exploration. 
As such, we extend the BCQ algorithm by adding a learnable parameter noise \cite{fortunato2017noisy} to the perturbation model $\xi_{\phi} (s, a)$ as $\xi_{\phi'} (s, a)$. Taking a fully-connected layer $y=wx$ as an example, where $x \in \mathbb{R}^p$ and $y \in \mathbb{R}^q$ are the input and output features, respectively, and $w \in \mathbb{R}^{q \times p}$ is the network parameter. Then the corresponding network with perturbation parameter noise is modified as:
\begin{equation}
y = w'x = (\mu^w + \sigma^w \cdot \epsilon^w)x,
\label{networknoisy}
\end{equation}
where the parameters $\mu^w \in \mathbb{R}^{q \times p}$ and $\sigma^w \in \mathbb{R}^{q \times p}$ are learnable parameters of the perturbation network. Here, $\epsilon^w \in \mathbb{R}^{q \times p}$ are noisy random variables that can be learned through back-propagation. The modified perturbation model is thus updated as: 
\begin{equation}
\label{eqn:perturb}
\phi' \leftarrow \argmax_{\phi'} \sum_{(s, a) \in \mathcal{B}} Q_{\theta} (s, a^n_i + \xi_{\phi'} (s, a^n_i)),
\end{equation}
where $\phi'$ is the parameter of the new perturbation model after incorporating the learnable noise parameters. 

\subsection{Learning to Provide Safety Guarantee}
We consider the case that the operation space is defined and restricted based upon those observed within the static dataset $\mathcal{B}$. We aim at enhancing the BCQ algorithm with guaranteed safety. Towards that end, we perform a joint learning framework to obtain the system dynamics in Eqn.~\ref{equ:state_Trans} together with its Lyapunov function. This collective learning schemes ensures system stability according to the Lyapunov stability criterion introduced in Section~\ref{subsec:LyapStab}. Specifically, we define a ``nominal'' closed-loop system dynamics $\bar{f}_{\psi_1}(\cdot)$ and the corresponding Lyapunov function $V_{\psi_2}(\cdot)$ as two neural networks. From \cite{manek2020learning}, it follows that: 
\begin{equation}
    f_{\psi_1}(s)=\bar{f}_{\psi_1}(s)-\nabla_{f_{\psi_1}}V_{\psi_2}(s)\frac{\sigma\big(\nabla_{f_{\psi_1}}V_{\psi_2}(s)^\top\bar{f}_{\psi_1}(s)+\alpha V_{\psi_2}(s)\big)}{||\nabla_{f_{\psi_1}}V_{\psi_2}(s)||_2^2},
\end{equation}
where the structure of $\bar{f}_{\psi_1}$ can be conveniently chosen as random fully connected network whereas the network for Lyapunov function learning is generally chosen as Input Convex Neural Network (ICNN) \cite{amos2017input}. Here $\alpha$ is an assigned parameter, and $\sigma(\cdot)$ is a smoothed ReLU activation with a quadratic region in $[0,l]$:
\begin{equation}
    \sigma(x)=
    \begin{cases}
    0, & \text{if } x\leq0\\
    \frac{x^2}{2l}, & \text{if } 0<x<l\\
    x-\frac{l}{2}, & \text{otherwise}.\\
    \end{cases}
\end{equation}

By enforcing that no positive component of $\nabla_{f_{\psi_1}}V_{\psi_2}(s)$ is along  the direction of $f_{\psi_1}(s)$, according to the afore-mentioned Lyapunov stability theory, the stability of $f_{\psi_1}(s)$ is guaranteed.

Furthermore, in addition to system stability, we also seek to provide safety guarantees with the optimized solution from the exploration policy. According to Eqn.~\ref{lf criteria}, an extended Lyapunov function design can be formulated as the following mini-max based cost function \cite{chang2019neural}:
\begin{equation}
\inf_{\psi_1} \sup_{s \in \mathcal{B} } \left[ \max(0,-V_{\psi_2}(s))+\max (0,\nabla_{f_{\psi_1}}V_{\psi_2}(s))+V_{\psi_2}^2(0)) \right].
\label{lfcost}
\end{equation}

Note that even the convexity of ICNN ensures that $V(\cdot)$ has only a single global optimum \cite{amos2017input}, it does not require the optimum is at $s=0$. To address this issue while avoiding increased computational burden and maintaining the function convexity, we perform an internal kernel function shifting \cite{manek2020learning} to achieve $V(0)=0$. In the meantime,  a small positive term is added to ensure strict positive-definiteness:
\begin{equation}
V_{\psi_2}(s) = \sigma(g(s)-g(0)) + \epsilon||s||^2,
\end{equation}
where $\epsilon$ is a small constant and $g(\cdot)$ is an ICNN. In practice, Eqn.~\ref{lfcost} can be solved as the following empirical Lyapunov risk index through Monte Carlo estimation,
\begin{equation}
\resizebox{.70\hsize}{!}{${L_s} = E_{s \sim \rho (\mathcal{B})} \lbrack \max(0,-V_{\psi_2}(s)) + \max (0,\nabla_{f_{\psi_1}}V_{\psi_2}(s))+V_{\psi_2}^2(0)) \rbrack$},%
\label{lfrisk}
\end{equation}
where $s$ is the state variable sampled according to distribution $\rho$ from the data batch $\mathcal{B}$. 
Finally, the following Lyapunov risk is added to the critic network as:
\begin{equation}
\theta, \psi_1, \psi_2 \leftarrow \argmin_{\theta, \psi} (\sum_{(s, a) \in \mathcal{B}} (y - Q_{\theta} (s, a)) + L_s).
\end{equation}

Pseudo-code of the proposed offline RL algorithm with enhanced safety and promoted exploration  is summarized in Algorithm~\ref{algorithm1}, and the major changes from the BCQ algorithm are highlighted in blue.

\begin{algorithm}[H]
\caption{Improved BCQ with safety and exploration enhancement}\label{euclid}
\begin{algorithmic}[1]
\STATE \textbf{Input:} Batch of data $\mathcal{B}$, horizon $T$, target network update rate $\tau$, mini-batch size N, number of sampled action $n$,
minimum weighting $\lambda$.\\
Initialize Q-networks $Q_{\theta_1}$ and  $Q_{\theta_2}$, \textcolor{blue}{noisy perturbation network $\epsilon_{\phi'}$}, VAE $G_w=(E_{w_1},D_{w_2})$ and \textcolor{blue}{Lyapunov function $V_{\psi}$}, with random parameters $\theta_1, \theta_2, \phi, w$ and target network $Q_{\theta_1'}$ and $Q_{\theta_2'}$ with $\theta_1' \leftarrow \theta_1$, $\theta_2' \leftarrow \theta_2$.
\FOR {episode $t=1$ to $T$ do}
\STATE Sample mini-batch of N transitions $(s,a,r,s')$ from $\mathcal{B}$
\STATE $\mu,\sigma= E_{w1}(s,a)$, $\dot{a}=D_{w_2}(s,z)$, $z \sim \mathcal{N}(\mu,\sigma)$
\STATE w $\leftarrow$ $\argmin_w \sum (a-\dot{a})^2 + D_{KL}(\mathcal{N}(\mu,\sigma)||\mathcal{N}(0,1))$ 
\STATE Sample $n$ actions: ${a_i \sim G_w(s')}^n_{i=1}$
\STATE \textcolor{blue}{(Explore efficiently) Generate perturbed actions: ${a_i \sim a_i +  \epsilon_\phi'(s')}^n_{i=1}$}
\STATE \textcolor{blue}{(Guarantee Safety) Compute Lyapunov risk $L_s$ according to Eq.\ref{lfrisk}}
\STATE Compute value target $y$ (Eqn.~\ref{eqn:target})
\STATE \textcolor{blue}{$\theta, \psi \leftarrow \argmin_{\theta, \psi} (\sum_{(s, a) \in \mathcal{B}} (y - Q_{\theta} (s, a)) + L_s)$}
\STATE \textcolor{blue}{$\phi' \leftarrow \argmax_{\phi'} \sum_{(s, a) \in \mathcal{B}} Q_{\theta} (s, a^n_i + \xi_{\phi'} (s, a^n_i))$}
\STATE Update target network: $\theta_i' \leftarrow \theta_i$
\ENDFOR
\\ {*} The major changes from the BCQ algorithm are highlighted in blue.
\end{algorithmic}
\label{algorithm1}
\end{algorithm}

\section{Experiments}
\label{sec:exp}

\subsection{Experimental Setup}
We apply our new offline RL framework to autonomous driving tasks, where  the open-sourced gym-based  environment, highway-env simulator\footnote{\url{https://highway-env.readthedocs.io/en/latest/}}, is adapted as our simulation platform. In this platform,  vehicle trajectories  are generated based on the kinematic bicycle model \cite{polack2017kinematic}, where the vehicles take continuous-valued actions for steering and throttle controls as defined in \cite{highway-env}. To collect data for offline RL training, a DDPG agent over 5,000 time steps is trained and the experience buffer $\mathcal{B}$ is trained. We use the DDPG implementation from the OpenAI baselines\footnote{\url{https://stable-baselines.readthedocs.io/en/master/}}. The proposed approach is experimented on the following two traffic scenarios.

\subsubsection{Highway scenario}
 The highway environment is illustrated in Fig.~\ref{fig:lane_change_construction}, where autonomous vehicle (AV, blue) intends to navigate as fast as possible without colliding with the human-driven vehicles (HDVs, green). The AV is expected to make lane changes to overtake slow-moving vehicles whenever possible to achieve higher speed. 
The reward function is defined as:
\begin{equation}
r = \alpha \times \frac{v-v_{min}}{v_{max}-v_{min}} - \beta \times collision,
\label{highway}
\end{equation}	
where $v, v_{min}, v_{max}$ are the current, minimum and maximum speed of the ego-vehicle, respectively, and $\alpha=0.4\text{ and } \beta=1$ are two weighting coefficients. 

\subsubsection{Parking scenario}
\begin{figure*}[!ht]
    \centering
    \subfloat[\centering Lane-change scenario. \label{fig:lane_change_construction} ]{{\includegraphics[width=.49\linewidth]{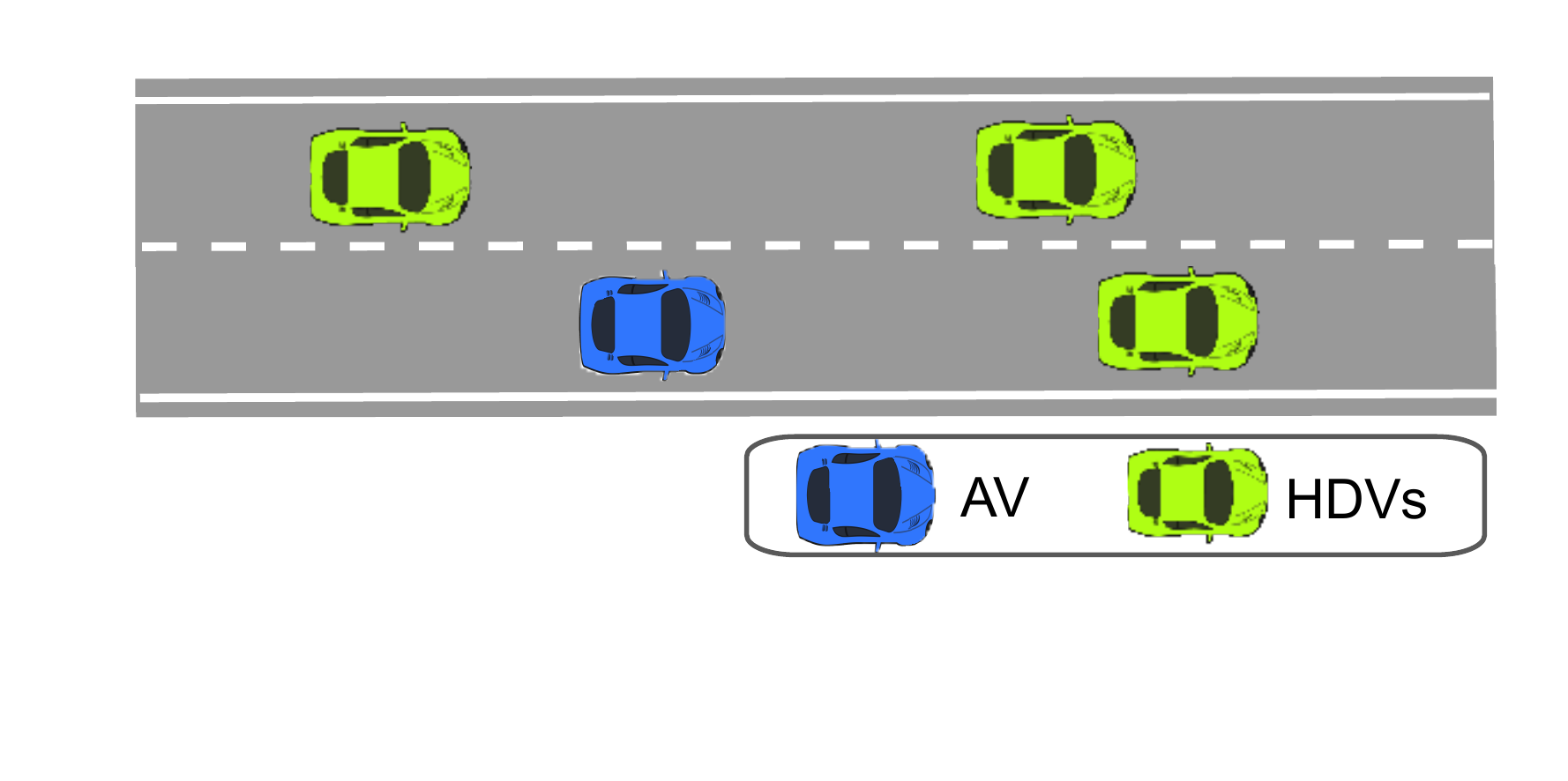} }}%
    \qquad
    \subfloat[\centering Parking scenario. \label{fig:parking-sce} ]{{\includegraphics[width=.37\linewidth]{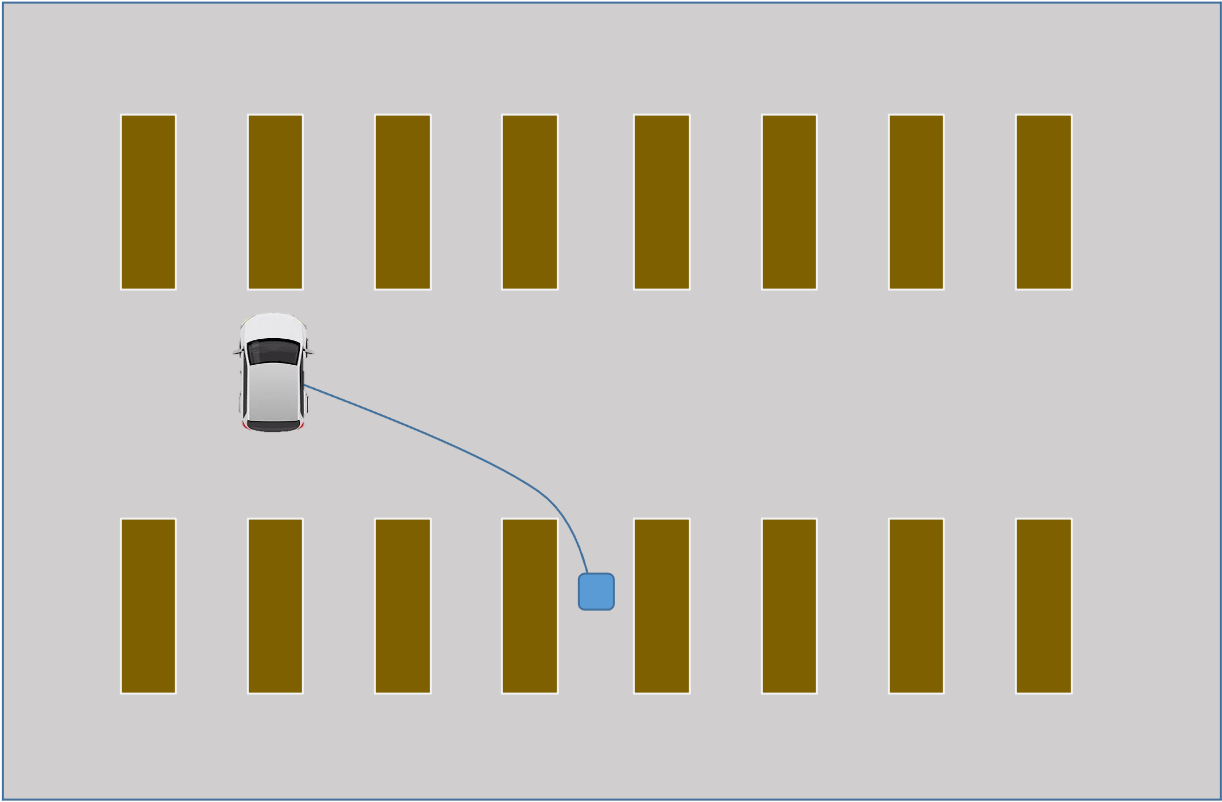}}}
    \caption{Two traffic scenarios: freeway and parking.}
    \label{fig:scenarios}
    % \vspace{-15pt}
\end{figure*}

Fig.~\ref{fig:parking-sce} shows the parking scenario, where the objective of the AV is to park successfully to stay within a desired space with appropriate heading while not colliding with the obstacles (dark green boxes). In this scenario, the reward is defined as:
\begin{equation}
r = -\alpha \times ||s-s_g||^2 - \beta \times violation,
\label{highway}
\end{equation}	
where $s = [x, y, v_x, v_y, cos(\phi), sin(\phi)]$ represents the current state of the AV whereas  $s_g = [x_g, y_g, 0, 0, cos(\phi_g), sin(\phi_g)]$ is the goal position and orientation.  The violation term represents the penalty on hitting obstacles. Here $\phi$ is the heading angle, and $\alpha=1 \text{ and } \beta=5$ are two weighting coefficients. 

\subsection{Baselines}
To demonstrate the effectiveness of our proposed approach, we compare our approach with a state-of-the-art conventional off-policy RL, as well as BCQ, a state-of-the-art offline RL algorithm:
\begin{enumerate}
\item \textit{Deep Deterministic Policy Gradient (DDPG)} \cite{lillicrap2015continuous}: DDPG is an off-policy deterministic version  of model-free RL algorithm that can handle continuous action space. We adapt the implementation based on OpenAI stable baseline\footnote{\url{https://stable-baselines.readthedocs.io/en/master/}}.
\item \textit{Batch Constraint Reinforcement Learning (BCQ)} \cite{fujimoto2019benchmarking}: BCQ is a state-of-the-art offline RL algorithm for continuous control with a state-dependent generative model used to restrict predicted actions to be similar to previous observed ones.
\item \textit{Noisy BCQ}: In this version, we extend BCQ by only adding the exploration-promotion strategy on the policy as detailed in Section~\ref{sec:noi}, without employing any safety-enhancement schemes.  
\item \textit{Ours}: The framework extends BCQ by incorporating  a new perturbation model with learnable parameter noise as well as a Lyapunov-based  safety-enhancement scheme.
\end{enumerate}
For this comparsion, we train all algorithms over 200 episodes and evaluate the models every 10 episodes with 5 different random seeds while the same random seeds are shared among the models. We set the discount factor $\gamma$ as 0.7.

\subsection{Performance Comparison}
\subsubsection{Comparison with state-of-the-art benchmarks}
\begin{figure*}[!ht]
    \centering
    \subfloat[\centering Returns in highway. \label{fig:highway} ]{{\includegraphics[width=.45\linewidth]{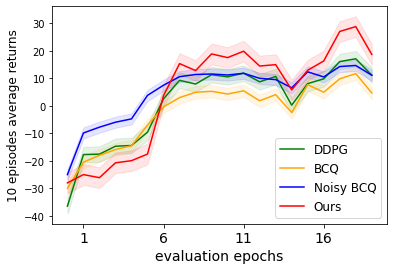} }}%
    \qquad
    \subfloat[\centering Returns in parking. \label{fig:parking} ]{{\includegraphics[width=.45\linewidth]{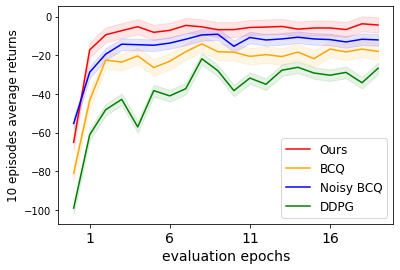}}}
    \caption{Comparison on evaluation returns between the proposed approach and state-of-the-art benchmarks.}
    \label{fig:returns}
    % \vspace{-15pt}
\end{figure*}

The comparison between  the proposed algorithm and state-of-the-art off-policy and offline algorithms are shown in Fig.~\ref{fig:highway} and Fig.~\ref{fig:parking} on the highway and parking scenarios, respectively. It is clear that our proposed approach consistently outperforms the benchmark algorithms in terms of evaluation returns and training efficiency, which is a result of the proposed parameter noise injection and safety guarantee schemes that facilitate exploration and enhance system safety.  It is also noted that Noisy BCQ also outperforms standard BCQ  in both traffic scenarios, which demonstrates that adding parameter noises to the perturbation model in BCQ can promote efficient explorations in BCQ. 

\begin{figure*}[!ht]
\centering
\includegraphics[width=1.02\textwidth]{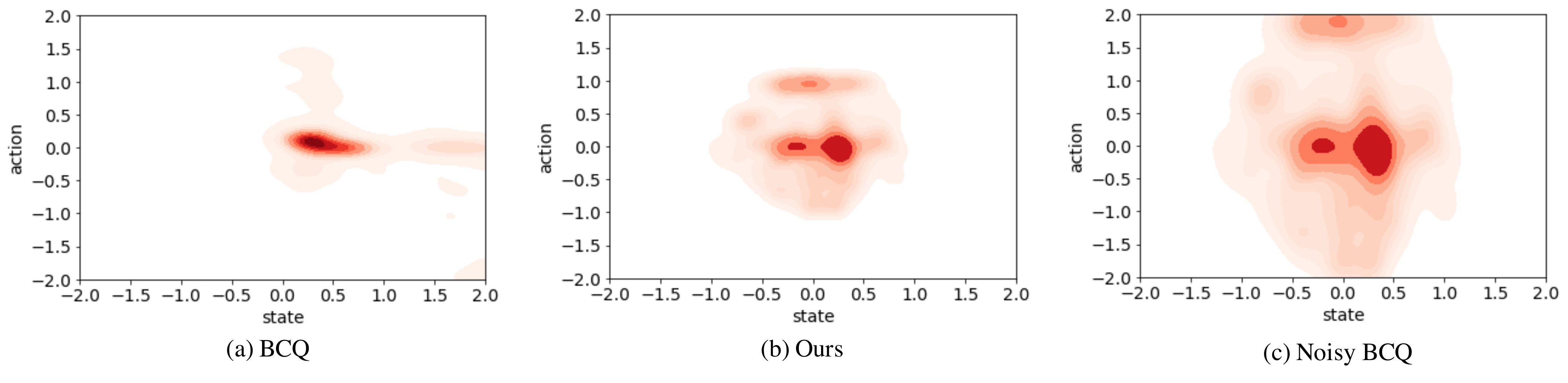}
\caption{State action density contours in the parking scenario. Darker colors represent more frequent state-action pairs.}
\label{density}
\end{figure*}

To show the correlation between state and action pairs, we plot the state-action density  in the parking scenario  in Fig.~\ref{density}, where we transform the multi-dimensional features of state and action into one dimensional vectors using principal component analysis (PCA) to show the diversity of the observed state-action pairs. It can be seen that BCQ explores rather ``cautiously'' 
with very limited state and action space. 
In contrast, the Noisy BCQ exhibits more efficient and ``aggressive'' exploration, surveying a  much larger state-action space. This demonstrates that the proposed parameter noise injection scheme can effectively promote exploration in  BCQ. With additional Lyapunov-based safety-enhancement, our proposed approach shows the same range of visited action space as Noisy BCQ but restricts the state space in a reasonable range, striking a good balance between exploration and safety as can be seen next.
     
\subsubsection{Performance of safety enhancement}
\begin{figure*}[!ht]
    \centering
    \subfloat[\centering Minimum distance with our approach. \label{fig:our-dis} ]{{\includegraphics[width=.45\linewidth]{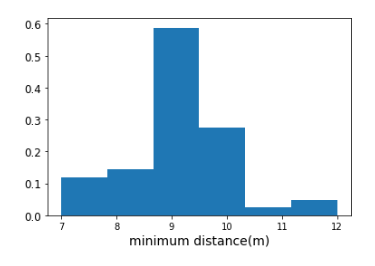} }}%
    \qquad
    \subfloat[\centering Minimum distance with noisy BCQ. \label{fig:noisybcq-dis} ]{{\includegraphics[width=.45\linewidth]{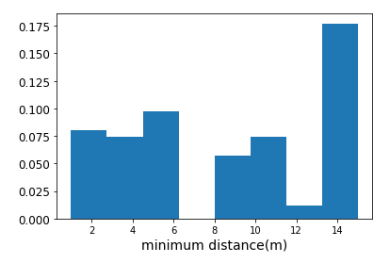}}}
    \caption{Comparison on minimum distance between our method and Noisy BCQ.}
    \label{fig:dis}
    % \vspace{-15pt}
\end{figure*}

To evaluate the performance of the proposed safety scheme, we compare the proposed approach with the Noisy BCQ that only has the parameter noise injection scheme without safety enhancement. Fig.~\ref{fig:dis} shows the minimum distance to the surrounding vehicles in the highway scenario for the proposed approach and Noisy BCQ. It is obvious that our approach presents a much higher minimum distance than Noisy BCQ which frequently leads to distances smaller than 5 $m$. This is because Noisy BCQ only promotes exploration without considering the safety issues.

\begin{figure*}[!ht]
    \centering
    \subfloat[\centering Comparison on steering performance. \label{fig:steer} ]{{\includegraphics[width=.45\linewidth]{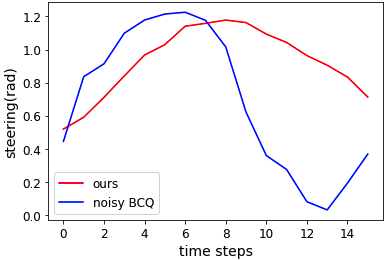} }}%
    \qquad
    \subfloat[\centering Comparison on acceleration performance. \label{fig:acc} ]{{\includegraphics[width=.45\linewidth]{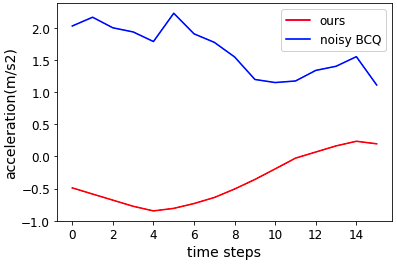}}}
    \caption{Performance comparison on steering and acceleration.}
    \label{fig:steeing_acc}
    % \vspace{-15pt}
\end{figure*}

\begin{figure}[!ht]
	\centerline{\includegraphics[width=0.48\textwidth]{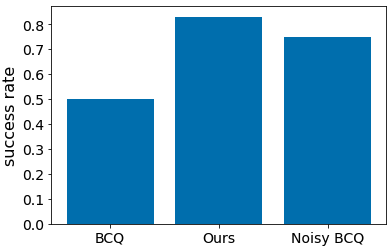}}
	\caption{Comparison on different success rates in the parking scenario.}
	\label{sr}
\end{figure}

Furthermore, we compare the performance of our approach with Noisy BCQ in the parking scenario in terms of steering angle, acceleration and success rat. As shown in Fig.~\ref{fig:steer}, our proposed approach has a smooth steering angle than Noisy BCQ which has sharp changes in steering angle that is risky and leads to poor ride comfort in real-world driving. The acceleration plots in Fig.~\ref{fig:acc} indicates that our approach also has a lower acceleration compared to Noisy BCQ. Higher and more oscillatory accelerations can cause very poor drive comfort and reduce the lifespan of vehicles. Above all, our approach achieves the highest success rates than the BCQ and Noisy BCQ in the parking scenario  as shown in Fig.~\ref{sr}.

\section{Conclusion and Future Work}
\label{sec:conclu}
In this paper, we developed  an efficient and safety-enhanced offline RL framework with application to autonomous driving in highway and parking traffic scenarios. To facilitate  exploration, we improved the BCQ algorithm by exploiting learnable parameterized noises in the perturbation model. A novel safety scheme was developed using Lyapunov stability theory to enhance  safety during explorations. Comprehensive experiments on the application of autonomous driving were conducted to compare our approach with several state-of-the-art algorithms, which demonstrated that the proposed approach consistently outperformed the benchmark approaches in terms of training efficiency and safety. In our future work, we plan to collect and employ more diverse data such as data from conventional control methods and real-world data from autonomous vehicles to further improve the performance.

\small{
\bibliographystyle{IEEEtran}
\bibliography{mylib.bib} }

\end{document}